\documentclass[letterpaper]{article} 
\usepackage{aaai2027}  
\usepackage[hyphens]{url}  
\usepackage{graphicx} 
\usepackage{natbib}  
\usepackage{caption} 
\usepackage{algorithm}
\usepackage{algorithmic}
\usepackage{amsmath}
\usepackage{booktabs}
\usepackage{multirow}
\usepackage{subcaption}
\usepackage{graphicx}
\usepackage{amssymb}

\usepackage{newfloat}
\usepackage{listings}
\DeclareCaptionStyle{ruled}{labelfont=normalfont,labelsep=colon,strut=off} 
\floatstyle{ruled}
\newfloat{listing}{tb}{lst}{}
\floatname{listing}{Listing}

\usepackage{booktabs}

\title{Geometry of Divergence: \\ Tracking Hidden-State Trajectories for Adaptive Multi-Turn Reasoning}

\author{
    Jie Liang\textsuperscript{\rm 1},
    Zhengxin Yu\textsuperscript{\rm 1}\corresponding,
    Hamid Nasiri\textsuperscript{\rm 2},
    Peter Garraghan\textsuperscript{\rm 1}
}
\affiliations{
    \textsuperscript{\rm 1}School of Computing and Communications, Lancaster University, UK\\
    \textsuperscript{\rm 2}School of Computing and Engineering, University of Huddersfield, UK\\
    \{j.liang14, z.yu8, p.garraghan\}@lancaster.ac.uk, h.nasiri@hud.ac.uk
}

\nocopyright

\begin{document}

\maketitle

\begin{abstract}
LLM agents need to sustain goal-consistent reasoning across long multi-turn interactions under strict resource constraints. 
However, as the multi-turn context accumulates, it can destabilize the underlying LLM's internal representation of task-relevant information from earlier turns, blurring the boundary between constructive reasoning and representation drift. 
We formulate multi-turn reasoning as a hidden-state trajectory of the underlying LLM that is characterized via two complementary signals: temporal curvature that captures the directional consistency of turn-to-turn updates, and variance slope which measures the expansion or contraction of the exploration space. 
Across four tasks and three underlying LLMs, we observed that these geometric signals distinguish between correct and incorrect episodes prior to completion. 
We further decompose each episode into three-action chains formed from four actions (Read, Write, Respond, Transfer) and show that separability is action-dependent, with different signals distinguishing various chain patterns. 
Our experiments demonstrate that trajectory geometry can identify critical turns in the reasoning process, increasing task success rates on $\tau$-Bench from 24.1\% to 39.6\% while reducing token cost by 11.2\%. 

\end{abstract}



\section{Introduction}

Recent advances in Large Language Models (LLMs) have accelerated a paradigm shift toward autonomous LLM agents that commonly operate through multi-turn interactions. Rather than fully specifying their task objectives at the outset, users refine their requirements over multiple turns by adding details and constraints based on the agent's intermediate feedback  \cite{laban2025llms}. Multi-turn interaction can therefore be viewed as an incremental reasoning process that reduces epistemic uncertainty about the task. In resource-constrained settings (\textit{e.g.,} limited model size and finite context windows), accumulating interaction history can lead to context saturation and reduce attention efficiency \cite{liu2024lost}.
Maintaining long-horizon, goal-consistent reasoning under such resource constraints has emerged as a fundamental challenge in LLM agent research.


\begin{figure}[t]
    \centering
    \includegraphics[width=0.9\linewidth]{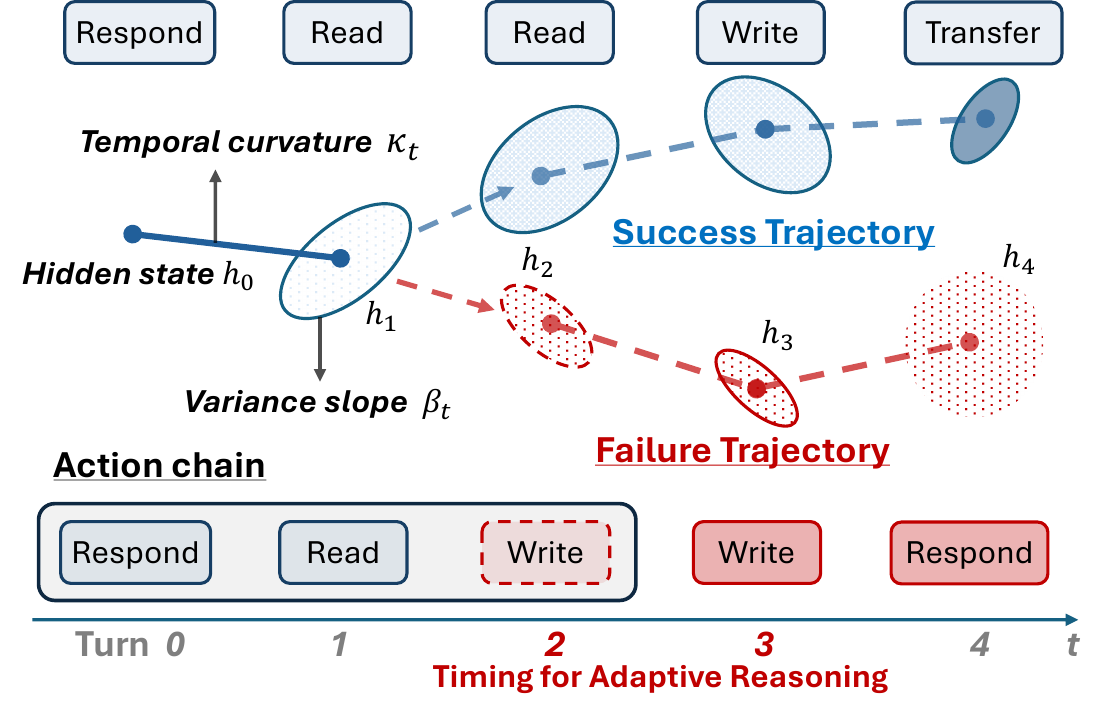}
    \caption{Multi-turn reasoning as an observable hidden-state trajectory.}
    \label{fig:overview}
\end{figure}

Existing approaches primarily rely on explicit experience-based heuristic methods of context engineering such as planning, reflection loop, and memory architecture to constrain the agent's inference trajectory and improve reasoning over long multi-turn interactions \cite{mei2025survey}.
However, these methods do not resolve the mismatch between agent goals and the utilized LLM's objective. Agents are designed to complete tasks through multi-turn interactions, whereas their underlying LLMs are optimized to generate the best possible response at each turn.
Existing Large Reasoning Models (LRMs) also use a fixed level of reasoning effort during dialogue, without adapting it to the needs of each turn \cite{zhao2026trade}. Unnecessary reasoning can introduce semantic noise and irrelevant counterfactuals into the autoregressive history, corrupting the agent’s internal representation of accumulated constraints.

Diagnosing and mitigating this internal corruption requires tracking how the underlying LLM's hidden states change across turns. Capturing the hidden-state trajectory is more difficult in multi-turn interactions compared to single-turn or basic Chain-of-Thought (CoT) reasoning \cite{wang2025latent}. Each hidden state update reflects both current task reasoning and the accumulated influence of earlier turns. This makes it difficult to distinguish useful hidden state updates from representation drift, and to determine when intervention is needed. Therefore, identifying the precise inflection points at which hidden states begin to deviate from the core objective is key to engineering reliable agents.



To address this challenge, we formulate multi-turn reasoning as an observable hidden-state trajectory of the underlying LLM and, as illustrated in Fig.\ref{fig:overview}, track where correct and incorrect episodes diverge in order to decide when to intervene, triggering adaptive reasoning precisely at the turns where failure is imminent. Our contributions are threefold:

\begin{itemize}
\item We characterize the trajectory with two complementary geometric signals, temporal curvature and variance slope. Across four tasks spanning open-ended reasoning (Math, Code) and tool-grounded interaction (Airline, Retail), and three underlying LLMs (Qwen3-14B, Qwen3-32B and Llama-3.1-8B), failure trajectories exhibit directional reversals and premature convergence in their hidden states.

\item We probe the source of this divergence by decomposing episodes into three-action chains, and find that separability is action-dependent, with information-output chains carrying the strongest discriminative structure while information-acquisition chains carry little.

\item We further demonstrate that these geometric signals can drive online control. By triggering adaptive reasoning at the turns where failure is imminent, our geometry-conditioned policy raises task success rates on $\tau$-Bench from 24.1\% to 39.6\% while reducing token cost by 11.2\%.
\end{itemize}

\section{Related Work}

\subsection{Multi-Turn Degradation in LLM Agents}
LLMs that perform well in single-turn settings degrade when task specifications are revealed incrementally. Laban et al. \cite{laban2025llms} report an average performance drop of 39\% across six generation tasks, attributing it to premature answer attempts and over-reliance on earlier, potentially erroneous turns. Realistic agentic settings amplify this problem. On $\tau$-Bench \cite{yao2024tau} and its dual-control successor $\tau^2$-Bench \cite{barres2025tau2}, even frontier function-calling agents solve fewer than half of the tasks and behave inconsistently across trials. To sustain long-horizon coherence, prior work engineers the explicit context, \textit{e.g.}, through verbal self-reflection \cite{shinn2023reflexion}, OS-inspired memory management \cite{packer2023memgpt}, and the context-engineering toolbox surveyed by Mei et al. \cite{mei2025survey}. These approaches operate on the textual history itself. In contrast, we monitor the latent trajectory induced by that history and intervene only when its geometry indicates drift.

\subsection{Hidden States Monitoring}
LLM internal states contain information predictive of output quality. Early work demonstrated that models can assess the validity of their own claims \cite{kadavath2022language}. Truthfulness is also linearly decodable from hidden states, with simple probes distinguishing true from false statements \cite{azaria2023internal,marks2023geometry}. At inference time, shifting activations along truth-correlated directions can further improve factuality \cite{li2023inference}.

Building on these findings, internal representations have been widely used for hallucination detection through the covariance spectrum of response embeddings \cite{chen2024inside}, probes over generation-time hidden states \cite{duan2024llms}, and attention-map features \cite{chuang2024lookback}.  Semantic entropy offers a complementary sampling-based uncertainty signal \cite{farquhar2024detecting} that can be approximated efficiently from the hidden states of a single generation \cite{kossen2024semantic}. Orgad et al. \cite{orgad2025llms} show that truthfulness information is concentrated in specific tokens and predicts the type of upcoming error. However, these signals do not transfer universally across tasks, which is consistent with the task-dependent behavior we observe for the variance slope. 

Among these methods, the ICR Probe is most closely related to our work. It tracks the cross-layer dynamics of hidden-state updates for hallucination detection \cite{zhang2025icr}, but analyzes individual responses in isolation. We characterize hidden-state trajectories across turns using temporal curvature and variance slope, and use their geometry as an online control signal rather than a post-hoc detector.

\subsection{Adaptive Reasoning Control}

Extended chain-of-thought is not uniformly beneficial, as reasoning models overthink simple problems \cite{chen2024overthinking}. In agentic environments, excessive deliberation can also displace environment interaction and lower task success \cite{cuadron2025overthinking}. Recent methods address this problem by selecting between thinking and no-thinking modes using reinforcement learning \cite{zhang2025adaptthink,fang2026thinkless} or Pareto-optimal chain-of-thought triggering \cite{lou2025adacot}. These methods are derived from the input question in largely single-turn settings. In contrast, our framework conditions the trigger on the agent's internal state accumulated across turns.

\section{Hidden State Trajectory Formulation}

\subsection{Multi-Turn Reasoning in Agent Tasks}

We begin by formalizing multi-turn reasoning in realistic agent settings. A typical agent multi-turn interaction comprises three components: (1) a system prompt $c_{\mathrm{sp}}$, which encodes the task specification and agent behavioral constraints; (2) a sequence of user inputs $\mathcal{Q} = \{q_0, q_1, \ldots, q_{T-1}\}$, representing $T$ alternating turns of interaction; and (3) a sequence of model outputs $Y = \{y_1, y_2, \ldots, y_T\}$ generated by a language model parameterized by $\theta$, where each $y_{t+1}$ is the output produced at turn $t$ conditioned on the system prompt and all preceding turns. The generation is formalized as the conditional distribution

\begin{equation}
    P_\theta(Y \mid c_{\mathrm{sp}}, \mathcal{Q}) = \prod_{t=0}^{T-1} P_\theta(y_{t+1} \mid c_{\mathrm{sp}}, y_{\leq t}, q_{\leq t}).
\end{equation}


In agent settings, the agent can also interact with the external environment $\theta'$ throughout the interaction. At each turn $t$, alongside its user-facing output, the agent may issue a tool invocation $y_t'$, yielding a response $a_t = \theta'(y_t')$ that is appended to the context and conditions all subsequent turns. We denote the set of such external interactions by $\mathcal{A} = \{a_t = \theta'(y_t')\}$. The full generation process thus becomes

\begin{equation}
    P_\theta(Y \mid c_{\mathrm{sp}}, \mathcal{A}, \mathcal{Q}) = \prod_{t=0}^{T-1} P_\theta(y_{t+1} \mid c_{\mathrm{sp}}, a_{\leq t}, y_{\leq t}, q_{\leq t}).
\end{equation}

\subsection{Hidden States in Multi-Turn Reasoning}

For each turn $t$, let $h_t \in \mathbb{R}^d$ denote the language model's hidden states at the final token position of the user input $q_t$, taken from a fixed intermediate probe layer (layer 22 for Qwen3-14B, layer 42 for Qwen3-32B, and layer 16 for Llama-3.1-8B). Under causal attention with residual connections, this representation encodes the full contextual semantics up to turn $t$; its projection onto the vocabulary determines the next-token distribution. We collect these representations into the trajectory
\begin{equation}
    \mathcal{H} = \{h_0, h_1, \ldots, h_{T-1}\}.
\end{equation}
The inter-turn dynamics are modeled as a residual process over turns
\begin{equation}
h_t = h_{t-1} + \mathbf{v}_t, \quad t = 1, \ldots, T-1,
\end{equation}
where the displacement $\mathbf{v}_t$ represents the change in the representation of the hidden state after turn $t$ relative to the accumulated context of all preceding turns.

\subsection{Temporal Curvature}

To measure the directional consistency of semantic drift across consecutive turns, we define the temporal curvature $\kappa_t$ at turn $t$ from the current hidden state $h_t$ and its two predecessors $h_{t-1}, h_{t-2}$, as the cosine of the angle between the two consecutive displacement vectors
\begin{equation}
    \kappa_t
    = \frac{\mathbf{v}_t \cdot \mathbf{v}_{t-1}}
           {\|\mathbf{v}_t\|_2\,\|\mathbf{v}_{t-1}\|_2},
    \quad
    t = 2, \ldots, T-1,
\end{equation}
where $\mathbf{v}_t = h_t - h_{t-1}$ and $\mathbf{v}_{t-1} = h_{t-1} - h_{t-2}$, defined whenever $\mathbf{v}_t \neq \mathbf{0}$ and $\mathbf{v}_{t-1} \neq \mathbf{0}$. The measure takes values in $[-1, 1]$. Values near $1$ indicate that context accumulates in a consistent direction, values near $0$ indicate orthogonal, non-reversing updates, whereas values near $-1$ indicate an abrupt reversal of the internal trajectory.

\subsection{Variance Slope}

For a prefix $\{h_0, \ldots, h_\lambda\}$ containing $\lambda+1$ hidden states, we define its dispersion $V(\lambda)$ as the trace of the sample covariance, using an unbiased estimator
\begin{equation}
\begin{aligned}
V(\lambda)
    &= \operatorname{tr}\!\big(\operatorname{Cov}(h_0,\ldots,h_\lambda)\big) \\
    &= \frac{1}{\lambda}\sum_{i=0}^{\lambda}
       \big\| h_i - \overline{h}^{(\lambda)} \big\|_2^2,
    \quad \lambda \geq 1,
\end{aligned}
\end{equation}

\begin{equation}
    \overline{h}^{(\lambda)}
    = \frac{1}{\lambda+1}\sum_{i=0}^{\lambda} h_i,
\end{equation}
where $\operatorname{tr}(\operatorname{Cov}(\cdot))$ is the sum of per-coordinate variances, and the normalization by $\lambda$ (rather than $\lambda+1$) reflects the unbiased estimator over the $\lambda+1$ prefix points. Varying $\lambda$ from $1$ to $t$ gives a dispersion sequence $\{V(1), \ldots, V(t)\}$, and we define the variance slope $\beta_t$ as its ordinary least-squares slope against prefix length
\begin{equation}
    \beta_t
    = \frac{\sum_{\lambda=1}^{t}\big(\lambda - \bar{\lambda}\big)
            \big(V(\lambda) - \bar{V}\big)}
           {\sum_{\lambda=1}^{t}\big(\lambda - \bar{\lambda}\big)^2},
    \quad t = 2, \ldots, T-1,
\end{equation}
where $\bar{\lambda}$ and $\bar{V}$ are the means of the prefix lengths and dispersion values. A positive $\beta_t$ indicates that representations progressively spread out as turns accumulate, whereas a negative value indicates a contracting dispersion.

\section{Empirical Analysis}
\label{sec:analysis}

\subsection{Experimental Setup}

\paragraph{Benchmarks}
Our experiments comprise four tasks across two multi-turn interaction scenarios.
The Math and Code tasks are drawn from the sharded-prompt framework of \emph{Lost in Conversation}~\cite{laban2025llms}, which demonstrates how an incorrect early assumption can derail a multi-turn conversation, inducing context contamination and progressively degrading reasoning efficiency. 
The Airline and Retail tasks are taken from $\tau$-Bench~\cite{yao2024tau}, which simulates dynamic dialogue between a user and a LLM agent equipped with domain-specific API tools and policy guidelines. 
Together, these benchmarks cover both open-ended reasoning (Math, Code) and tool-grounded, policy-constrained interaction (Airline, Retail).

\paragraph{Models}
We employ three LLMs of different sizes as the core components of both the assistant agent and the user simulator: Qwen3-14B, Qwen3-32B~\cite{qwen3technicalreport}, both LRMs with a toggleable thinking mode from long-CoT training, and Llama-3.1-8B~\cite{grattafiori2024llama31}.

\paragraph{Metrics}
Both benchmarks provide ground-truth reference answers, from which we derive a binary correctness label for each individual task. Aggregating tasks yields the average reward, which we report at the domain level.
To characterize the overhead of multi-turn reasoning, we record the true token cost of each task through the vLLM interface, accounting for context, tool calls, and thinking overhead.
Significance is assessed with the two-sided Mann-Whitney U test (p-values).

\paragraph{Implementation Details}
We conduct inference for LLMs using vLLM on 8 NVIDIA L40S GPUs and 4 H200 GPUs. All LLMs’ decoding hyper-parameters and prompt formatting strictly follow their respective official configurations. Details on benchmark selection for each analysis and the full experimental setup are provided in the Appendix.

\subsection{Temporal Curvature Analysis}
\label{}

\begin{figure*}[ht]
  \centering
  \begin{subfigure}[b]{0.49\textwidth}
    \centering
    \includegraphics[width=1\linewidth]{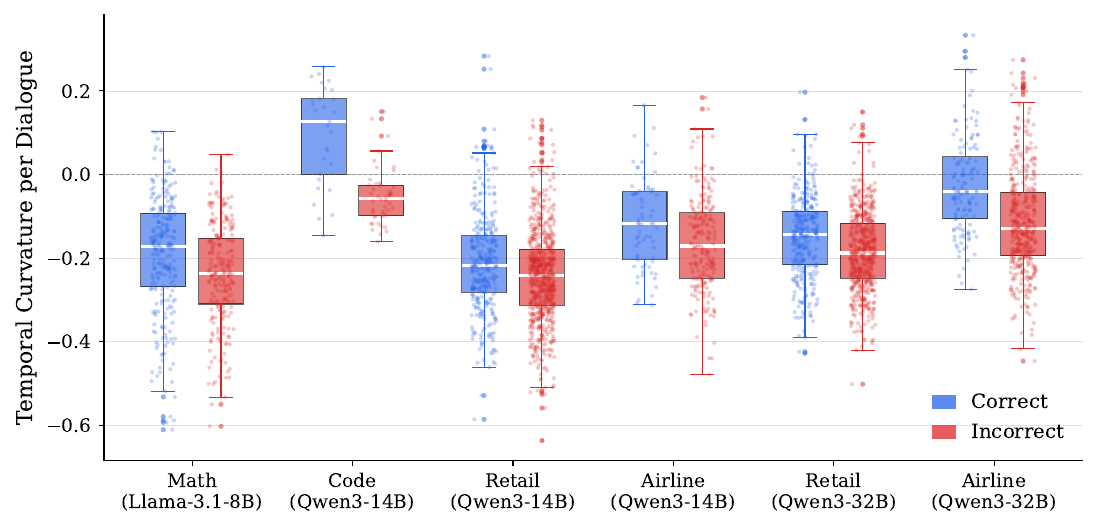}
    \caption{Temporal curvature $\kappa$: correct trajectories bend less.}
    \label{fig:curvature-box}
  \end{subfigure}
  \hfill
  \begin{subfigure}[b]{0.49\textwidth}
    \centering
    \includegraphics[width=1\linewidth]{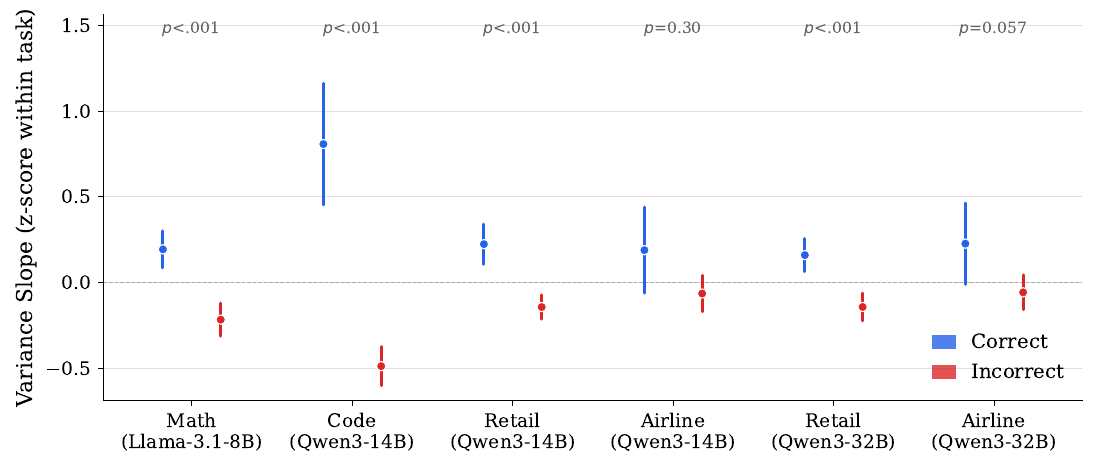}
    \caption{Variance slope $\beta_t$, z-scored within task.}
    \label{fig:varslope}
  \end{subfigure}
  \caption{Trajectory geometry separates correct from incorrect episodes across task and model settings.}
  \label{fig:geometry}
\end{figure*}

We compare temporal curvature $\kappa$ across four task types and three underlying LLMs. In Fig.\ref{fig:geometry}(a), the horizontal axis denotes each task-LLM pairing, and the vertical axis is the per-episode mean of $\kappa$. We make three main observations.

\paragraph{$\kappa$ medians are negative for most task–model pairings.} $\kappa$ characterizes the directional consistency of semantic drift between consecutive turns. A value of $\kappa<0$ indicates that a hidden-state update tends to reverse the previous turn's residual direction, which we interpret as consistent with a multi-turn reasoning process that produces an update that points in the opposite direction to the preceding update.
The Code correct group has a positive median (median = +0.127, mean = +0.096), suggesting that on structured, verifiable tasks hidden-state updates lean toward same-direction accumulation rather than reversal.

\paragraph{Failed episodes exhibit systematically more negative $\kappa$.} Across all six pairings, the mean of the failure group is consistently lower than that of the corresponding correct group, with differences (Correct$-$Incorrect) of: Math +0.047, Code +0.148, Retail (14B) +0.030, Airline (14B) +0.049, Retail (32B) +0.035, Airline (32B) +0.088. This indicates that failure trajectories undergo directional reversals in hidden-state updates more frequently, whereas correct trajectories have $\kappa$ closer to zero, and in some cases positive (\textit{e.g.,} the Code correct group), reflecting a tendency toward orthogonal or same-direction information accumulation.

\subsection{Variance Slope Analysis}
In Fig.\ref{fig:geometry}(b), the vertical axis is the variance slope $\beta_t$ standardized to zero mean and unit variance within each task. Recall that a positive $\beta_t$ indicates that the total dispersion of the hidden-state trajectory increases as the episode progresses, whereas a negative value indicates that it contracts.
\paragraph{Correct episodes exhibit higher variance slopes than incorrect episodes.} Across all six pairings, correct episodes have higher variance slopes than incorrect episodes. Correct trajectories maintain a broader exploration space across turns, whereas failed trajectories contract onto a narrower, unproductive region. The gap is largest on the sharded-framework tasks Math and Code and smallest on Airline. Separability is significant for all pairings except the two Airline cases (14B p=0.30, 32B p=0.057), tracking the increasing openness of the action space from structured tasks through Retail to Airline.

\subsection{Turn Alignment}
\begin{figure}[h]
\centering
\includegraphics[width=0.7\columnwidth]{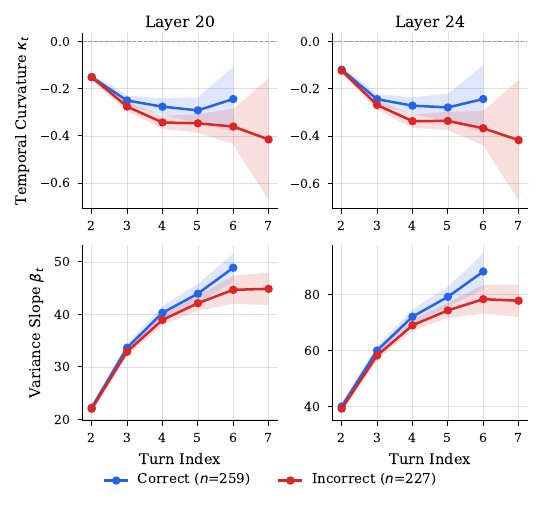}
\caption{Turn-resolved $\kappa_t$ and $\beta_t$ on Math tasks.}
\label{fig:combined-math}
\end{figure}

One methodological point deserves emphasis: all hidden states $h_t$ entering the curvature and variance-slope computations are turn-aligned. This alignment matters because trajectory length is a confound we observed directly, as incorrect episodes are systematically longer than successful ones due to retries, stalls, and back-and-forth entanglement, and a longer trajectory is mechanically more curved and accumulates more variance regardless of correctness. Indexing by turn removes this confound. Specifically, for the Math (as shown in Fig.\ref{fig:combined-math}), Code, and $\tau$-Bench datasets we index turns from 0 and take one $h_t$ per turn, the hidden state at the final token of the user input $q_t$ before the agent acts, then compute $\kappa$ and $\beta_t$ on the resulting sequence $\{h_0, h_1, \ldots\}$ of consecutive states. Because temporal curvature is a three-point statistic over $h_{t-2}, h_{t-1}, h_t$, values are undefined for the first two turns and available only from turn 2 onward, and we adopt the same alignment for the variance slope so the two metrics are defined on identical turn indices.

\begin{figure*}[t]
  \centering
  \begin{subfigure}[b]{0.32\textwidth}
    \centering
    \includegraphics[width=1\linewidth]{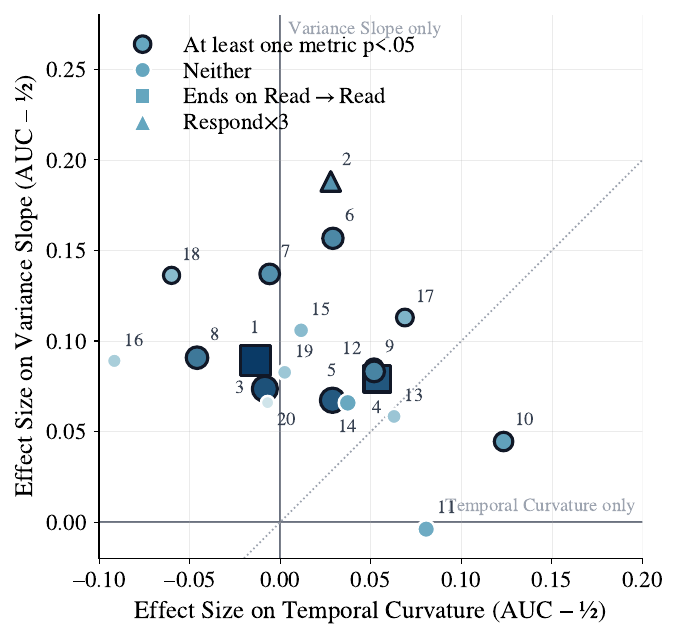}
    \caption{Effect sizes per action chain (Retail).}
    \label{fig:retail_effect_size}
  \end{subfigure}
  \hfill
  \begin{subfigure}[b]{0.32\textwidth}
    \centering
    \includegraphics[width=1\linewidth]{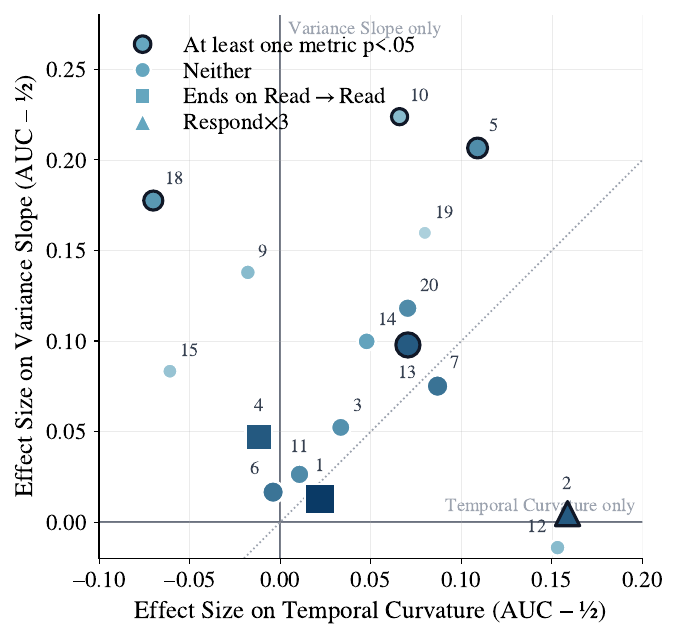}
    \caption{Effect sizes per action chain (Airline).}
    \label{fig:airline_effect_size}
  \end{subfigure}
  \hfill
  \begin{subfigure}[b]{0.32\textwidth}
    \centering
    \includegraphics[width=1\linewidth]{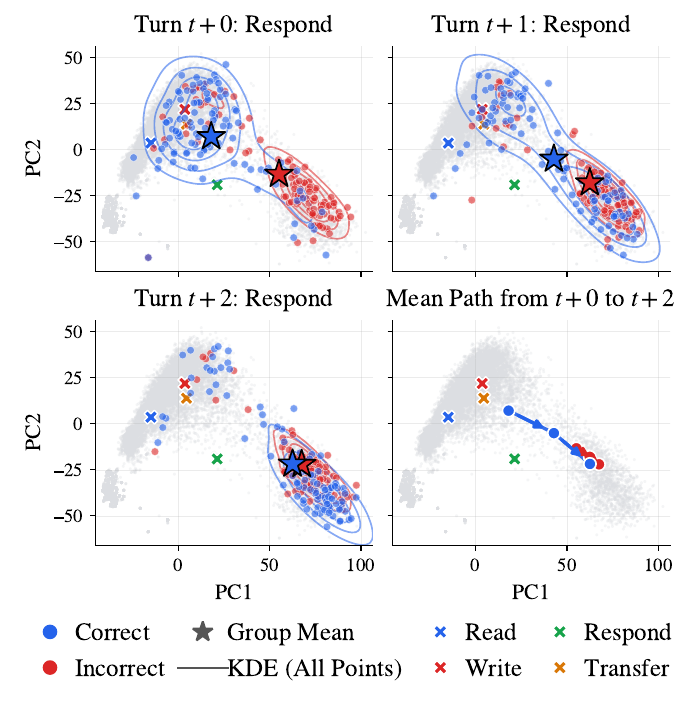}
    \caption{PCA trajectory of the Respond$\times 3$ chain.}
    \label{fig:respond_pca_trajectory}
  \end{subfigure}
  \caption{Discriminability of hidden-state trajectory geometry across action chains.}
  \label{fig:action_chains_in_geometry}
\end{figure*}

\section{Probing Trajectory Divergence through Action Chains}

\subsection{Action Chain Construction}

Based on the types of actions through which agents interact with the environment in $\tau$-Bench, we classify action chains into four categories: 
\begin{itemize}
\item \emph{Write} refers to write-type tool calls that alter the database state (\textit{e.g.,} update reservation, cancel pending order).
\item \emph{Read} refers to read-only tool calls that query the environment without altering its state (\textit{e.g.,} get reservation details, get user details).
\item \emph{Respond} refers to replies to the user simulator. 
\item \emph{Transfer} refers calls to \texttt{transfer\_to\_human}, which end the episode by escalating to a human agent.
\end{itemize}



We construct action chains by taking sliding windows of three consecutive turns over the recorded assistant actions. Table \ref{tab:chain_action_share} reports the proportion of three-action chains containing at least one such action in both the correct and incorrect episodes within each domain.

\begin{table}[h]
\centering
\small
\setlength{\tabcolsep}{4pt}
\renewcommand{\arraystretch}{1.15}
\caption{Share of three-action chains containing each action.}
\label{tab:chain_action_share}
\begin{tabular}{@{}lcccccc@{}}
\toprule
& \multicolumn{3}{c}{\textbf{Retail}} & \multicolumn{3}{c}{\textbf{Airline}} \\
\cmidrule(lr){2-4}\cmidrule(lr){5-7}
Action & Correct & Incorrect & $\Delta$ & Correct & Incorrect & $\Delta$ \\
\midrule
Write     & 49.3\% & 43.4\% & $-5.9$ & 15.3\% & 38.9\% & $+23.7$ \\
Read      & 85.8\% & 82.5\% & $-3.3$ & 86.1\% & 64.4\% & $-21.7$ \\
Transfer  & \phantom{0}4.2\% & \phantom{0}4.4\% & $+0.2$ & 29.2\% & \phantom{0}7.5\% & $-21.7$ \\
Respond   & 57.4\% & 50.0\% & $-7.4$ & 58.3\% & 61.7\% & $+3.4$ \\
\bottomrule
\end{tabular}
\end{table}

\begin{table}[h]
\centering
\footnotesize
\setlength{\tabcolsep}{4pt}
\renewcommand{\arraystretch}{1.05}
\caption{Top-20 three-action chains, with correct/incorrect window shares (\%) and per-domain frequency rank.}
\label{tab:top_chains}
\begin{tabular}{@{}clccccc@{}}
\toprule
& & \multicolumn{2}{c}{\textbf{Retail}} & \multicolumn{2}{c}{\textbf{Airline}} \\
\cmidrule(lr){3-4}\cmidrule(lr){5-6}
IDs & Action Chains & Cor/Inc & Rank & Cor/Inc & Rank \\
\midrule
1  & Read$\to$Read$\to$Read       & 19.6/22.2 & \#1  & 17.1/14.5 & \#2  \\
2  & Resp$\to$Resp$\to$Resp       & 3.6/8.2   & \#5  & 9.1/23.6  & \#1  \\
3  & Read$\to$Read$\to$Write      & 12.5/10.8 & \#2  & 3.2/8.0   & \#3  \\
4  & Resp$\to$Read$\to$Read       & 11.2/9.6  & \#3  & 9.3/5.1   & \#4  \\
5  & Read$\to$Write$\to$Resp      & 10.7/5.4  & \#4  & 3.4/4.5   & \#6  \\
6  & Read$\to$Read$\to$Resp       & 4.4/4.5   & \#6  & 5.5/4.7   & \#5  \\
7  & Read$\to$Resp$\to$Resp       & 3.7/2.6   & \#11 & 5.7/3.5   & \#7  \\
8  & Write$\to$Resp$\to$Resp      & 6.0/3.3   & \#7  & 0.7/2.8   & \#11 \\
9  & Read$\to$Write$\to$Read      & 2.4/4.0   & \#8  & 1.1/3.0   & \#9  \\
10 & Read$\to$Write$\to$Write     & 2.7/3.8   & \#9  & 1.1/2.7   & \#12 \\
11 & Read$\to$Resp$\to$Read       & 2.1/3.2   & \#12 & 3.6/2.4   & \#10 \\
12 & Resp$\to$Read$\to$Write      & 4.8/2.3   & \#10 & 1.1/1.9   & \#13 \\
13 & Read$\to$Read$\to$Transfer   & 0.9/1.4   & \#18 & 12.6/1.7  & \#8  \\
14 & Resp$\to$Read$\to$Resp       & 2.4/2.1   & \#13 & 2.2/1.5   & \#15 \\
15 & Write$\to$Read$\to$Write     & 1.2/2.0   & \#14 & 0.8/1.3   & \#18 \\
16 & Write$\to$Write$\to$Write    & 0.7/1.8   & \#16 & 0.3/1.4   & \#19 \\
17 & Write$\to$Write$\to$Resp     & 1.6/1.5   & \#15 & 0.5/1.2   & \#21 \\
18 & Resp$\to$Resp$\to$Read       & 1.2/1.5   & \#17 & 2.8/0.8   & \#20 \\
19 & Read$\to$Resp$\to$Write      & 0.8/1.0   & \#20 & 0.5/1.7   & \#16 \\
20 & Read$\to$Resp$\to$Transfer   & 0.3/0.5   & \#26 & 5.9/0.7   & \#17 \\
\bottomrule
\end{tabular}
\end{table}

\subsection{Separability of Hidden-state Trajectories in Action Chains}

\begin{center}
\fbox{
\begin{minipage}{0.95\linewidth}
Hidden-state trajectory separability varies with the distribution of agent actions across tasks.
\end{minipage}
}
\end{center}

\begin{figure*}[ht]
  \centering
  \begin{subfigure}[b]{0.49\textwidth}
    \centering
    \includegraphics[width=0.9\linewidth]{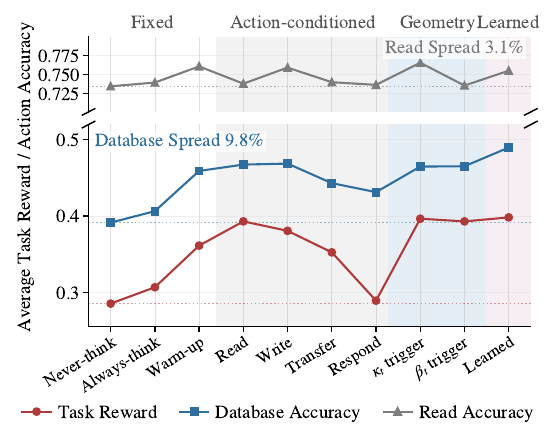}
    \caption{Retail domain.}
    \label{fig:curvature-box}
  \end{subfigure}
  \hfill
  \begin{subfigure}[b]{0.49\textwidth}
    \centering
    \includegraphics[width=0.9\linewidth]{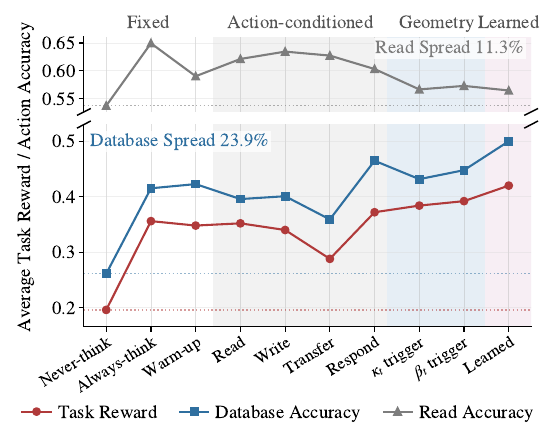}
    \caption{Airline domain.}
    \label{fig:varslope}
  \end{subfigure}
  \caption{Trigger-policy comparison on Qwen3-14B across the Retail and Airline domains.}
  \label{fig:comparison}
\end{figure*}

As shown in Table\ref{tab:chain_action_share} and Fig.\ref{fig:retail_effect_size} and Fig.\ref{fig:airline_effect_size}, the discriminative capabilities of the two probes exhibit systematic differences across the two task categories. In the Retail task, differences in domain action distributions between correct and incorrect episodes are relatively small. The discriminative signal primarily comes from the variance slope. Temporal curvature provides comparatively limited separation. Conversely, in the Airline task, differences in domain action distributions between correct and incorrect episodes are larger. Variance slope shows stronger discriminative power than in Retail, while temporal curvature also demonstrates marked discriminative capability. This comparison shows that temporal curvature and variance slope are not redundant probes, but rather provide complementary signals whose relative discriminative power varies across tasks.

\begin{center}
\fbox{
\begin{minipage}{0.95\linewidth}
Action chains dominated by information acquisition exhibit limited discriminative power (\textit{e.g.,} IDs 1 and 4), whereas those dominated by information output and action switching demonstrate significant discriminative power (\textit{e.g.,} IDs 2 and 10).
\end{minipage}
}
\end{center}

Classifying action chains by function: chains with IDs 1 and 4 (where the last two steps consist of consecutive Read actions, or where the chain as a whole is dominated by Read actions) represent behavior patterns dominated by information acquisition. These two types of chains account for a high proportion in both tasks, but their discriminative performance regarding correct versus incorrect is not particularly outstanding. By contrast, action chains dominated by information output and action switching, specifically IDs 2, 6, 10 and 17 in the Retail task, and IDs 2, 5, 10, 12, 19 and 20 in the Airline task that demonstrates significantly discriminative performance in both tasks.

\begin{center}
\fbox{
\begin{minipage}{0.95\linewidth}
Of all information-output chains, Respond$\to$Respond$\to$ Respond (ID 2) shows both the highest error concentration and the most pronounced discriminative structure in geometric terms.
\end{minipage}
}
\end{center}

In the geometric space shown in Fig.\ref{fig:retail_effect_size}, ID 2 (Respond$\to$Respond$\to$Respond) consistently corresponds to the most significant segmentation location. To further characterize its internal dynamics, we visualized the turn-wise trajectory of this chain in the two-dimensional PCA space corresponding to the hidden states (Fig.\ref{fig:respond_pca_trajectory}), and performed a numerical balancing of correct and incorrect samples. The results show that the scatter plot of correct episodes gradually shifts from the Write/Transfer region to the edge of the Respond cluster (bottom right corner) as the turns progress. In contrast, the scatter of incorrect episode remains in the bottom-right region of the Respond cluster throughout, showing no significant spatial expansion. This difference suggests that, in correct episodes, the latent state representation corresponding to the Respond action continuously expands outwards, whereas incorrect episodes mean remains comparatively localized in this PCA projection.

\section{Geometry-Triggered Adaptive Reasoning}
\label{sec:control}

\begin{table*}[ht]
\centering
\setlength{\tabcolsep}{5pt}
\caption{Trigger policies on $\tau$-Bench. The upper block reports task reward, the lower block mean token cost with the number of max-turn terminations in parentheses. \textbf{Bold} marks the best value in a row, \underline{underline} the runner-up.}
\renewcommand{\arraystretch}{1.15}
\begin{tabular}{ll ccc cc c}
\toprule
& & \multicolumn{3}{c}{\textit{Fixed policies}}
  & \multicolumn{2}{c}{\textit{Geometry-conditioned (training-free)}}
  & \textit{Learned} \\
\cmidrule(lr){3-5}\cmidrule(lr){6-7}\cmidrule(lr){8-8}
Domain & Model & Never-think & Warm-up & Always-think & Variance slope & Temporal curvature & Learned-HST \\
& & & & & $\beta_t>\tau_\beta$ & $\kappa_t<\tau_\kappa$ & \\
\midrule
\multicolumn{8}{l}{\textbf{Average task reward}}\\
Retail  & Qwen3-14B & 0.286 & 0.361 & 0.307 & 0.393 & \underline{0.397} \footnotesize($\tau_\kappa\!=\!-0.20$)\normalsize & \textbf{0.398} \\
Airline & Qwen3-14B & 0.196 & 0.348 & 0.356 & \underline{0.392} & 0.384 \footnotesize($-0.15$)\normalsize & \textbf{0.420} \\
Retail  & Qwen3-32B & 0.347 & 0.405 & 0.384 & \underline{0.412} & \textbf{0.442} \footnotesize($-0.15$)\normalsize & 0.405 \\
Airline & Qwen3-32B & 0.136 & 0.280 & \textbf{0.364} & 0.328 & \underline{0.352} \footnotesize($-0.10$)\normalsize & \textbf{0.364} \\
\midrule
\multicolumn{8}{l}{\textbf{Mean token cost} (thousands), with max-turn terminations}\\
Retail  & Qwen3-14B & 112.6 {\scriptsize(75)} & \underline{111.1} {\scriptsize(69)} & 113.1 {\scriptsize(100)} & 104.3 {\scriptsize(55)} & \textbf{101.0 {\scriptsize(34)}} & 104.4 {\scriptsize(51)} \\
Airline & Qwen3-14B & \phantom{0}82.5 {\scriptsize(28)} & \phantom{0}87.1 {\scriptsize(27)} & \phantom{0}83.5 {\scriptsize(26)} & \underline{\phantom{0}83.3} {\scriptsize(19)} & \textbf{\phantom{0}78.9 {\scriptsize(16)}} & \phantom{0}85.0 {\scriptsize(28)} \\
Retail  & Qwen3-32B & 125.7 {\scriptsize(117)} & 107.1 {\scriptsize(79)} & 119.3 {\scriptsize(108)} & \underline{105.1} {\scriptsize(70)} & \textbf{104.6 {\scriptsize(68)}} & 108.5 {\scriptsize(78)} \\
Airline & Qwen3-32B & \phantom{0}98.2 {\scriptsize(33)} & \phantom{0}92.6 {\scriptsize(35)} & 100.6 {\scriptsize(33)} & \underline{\phantom{0}86.3} {\scriptsize(25)} & \textbf{\phantom{0}83.0 {\scriptsize(25)}} & \phantom{0}87.5 {\scriptsize(29)} \\
\bottomrule
\end{tabular}
\label{tab:control}
\end{table*}

Sections 4 and 5 establish that hidden-state geometry separates correct from incorrect episodes prior to termination, with the effect most pronounced in chains dominated by information output. We now examine whether these signals can act as online control signals rather than post-hoc diagnostics, using the trajectory geometry to determine when the underlying LLM should engage its thinking mode.

We formalize every intervention as a binary decision $\delta_t \in \{0,1\}$ issued at each turn $t$, determining whether the agent enters thinking mode and generates a long-CoT trace before acting. Methods differ only in what information produces $\delta_t$:
\begin{itemize}
\item \textbf{Fixed Policies} set $\delta_t$ to a constant (Never-thinking, Always-thinking), and Warm-up,  which thinking only on the first two turns.
\item \textbf{Action-conditioned Policies} condition on a predicted next action $\hat{a}_t$ obtained from a learned classifier (F1: Read 0.826, Respond 0.711, Write 0.700, Transfer 0.531).
\item \textbf{Geometry-conditioned Policies} threshold the trajectory signals directly, firing when $\kappa_t < \tau_\kappa$ or $\beta_t > \tau_\beta$.
\item \textbf{Learned policy} trains a classifier on the concatenation of the recent three actions one-hots with hidden-state geometries $(\kappa_t, \beta_t)$, fitted on logged episodes to predict turns at which failure is imminent.
\end{itemize}

Because $\kappa_t$ is a three-point statistic, geometry-conditioned triggers are undefined before turn 2. To keep the comparison controlled, every method except Never-thinking adopts this same Warm-up.
Each configuration is run over 5 trials per task on the $\tau$-Bench Airline and Retail domains with Qwen3-14B and Qwen3-32B.

\paragraph{Deliberation gains depend on timing.} As shown in Fig.\ref{fig:comparison}, Always-thinking outperforms Never-thinking consistently, but the variation in gains does not support a simple ``more thinking is better'' interpretation. On Retail with Qwen3-14B, Always-thinking increases reward from 0.286 to 0.307, while Warm-up, which thinks on just the first two turns, reaches 0.361 at a lower token cost. On Qwen3-32B, Warm-up achieves 0.405 compared with 0.384 for Always-thinking, while using 12.2k fewer tokens. However, Warm-up does not perform consistently across settings. On Airline with Qwen3-32B, it reaches only 0.280 against Always-thinking's 0.364. These results indicate that task performance depends on the timing of deliberation, rather than its total amount alone.
 
\paragraph{The effectiveness of action-conditioned triggers varies across tasks.} Triggering on the appropriate action recovers much of the performance gain, but the most effective action does not transfer across domains. On Retail with Qwen3-14B, Read (0.393) and Write (0.381) perform well, while Respond collapses to 0.290. On Airline with the same model, the ordering inverts, with Respond (0.372) best and Read (0.352) and Write (0.340) both below Always-thinking. Therefore, the effectiveness of these triggers depends on the task-specific distribution of errors, which needs to be identified in advance.
 
\paragraph{Geometric thresholds are unimodal, transferable, and require no additional training.} Sweeping $\tau_\kappa$ is unimodal with an optimum locatable by a one-dimensional scan: on Retail with Qwen3-14B, thresholds of $-0.15$, $-0.20$, $-0.25$ give 0.388, 0.397, 0.363. That optimum shifts predictably with the correct$-$incorrect $\kappa$ gaps of Section~\ref{sec:analysis} (Retail 14B $+0.030$, Airline 14B $+0.049$, Retail 32B $+0.035$, Airline 32B $+0.088$), so it can be guided by the diagnostic statistics rather than searched per deployment. Crucially, the threshold matches or beats the trained classifier: $\kappa < -0.20$ ties Learned-HST on Retail 14B (0.397 vs 0.398) at lower token cost (101.0k vs 104.4k), and $\kappa < -0.15$ reaches 0.442 on Retail 32B, above the Learned-HST configuration there (0.405) and 27\% above Never-thinking. A trained trigger is fitted to one model's failure distribution, which moves with scale. Temporal curvature is distribution-free and adapts by recalibrating a single scalar.
 

\paragraph{Geometry-conditioned triggers combine token efficiency with competitive reward.}
Since token cost is influenced by both deliberation and failure length, firing at the wrong turns spends more while earning less. Across all four domain-model pairings, the geometry-conditioned triggers instead occupy the favorable corner of this trade-off, attaining the highest task reward among the lowest token costs, with max-turn terminations (default 50 turns) falling below every fixed policy (e.g., from 75 to 34 on Retail with Qwen3-14B under $\kappa<-0.20$). Averaging over the four settings, the geometry-conditioned trigger raises task reward from 0.241 under Never-thinking to 0.396, a gain from 24.1\% to 39.6\% in task success, while lowering mean token cost from 104.8k to 93.0k, an 11.2\% reduction.

\textbf{Applicability \& Future Direction.} We highlight two directions for future research.
First, trajectory geometry provides a new lens on multi-turn adversarial attacks, which is an active area in which jailbreaks are spread across turns to divert an agent away from its objective \cite{russinovich2025great}. Our signals could potentially detect such manipulation from the inside before it surfaces in the outputs. Second, as hidden states are increasingly used to detect and correct representation drift \cite{dongre2026attention}, characterizing trajectory geometry contributes an interpretable, training-free account of when drift begins, which is a step toward continually learning agentic systems that adapt reliably over long horizons.

\section{Conclusion}

In this work, we formulated multi-turn reasoning as an observable hidden-state trajectory of the underlying LLM and characterized it with two complementary geometric signals, temporal curvature $\kappa$ and variance slope $\beta$. Across four tasks and three underlying LLMs, these signals separate correct from incorrect episodes before termination, with failure trajectories exhibiting directional reversals and premature convergence in their hidden states. Decomposing episodes into three-action chains further showed that this separability is action-dependent, concentrated in information-output chains. Finally, we showed that trajectory geometry can act as an online control signal, triggering adaptive reasoning at critical turns, raises task success on $\tau$-Bench from 24.1\% to 39.6\% while reducing token cost by 11.2\%.

\bibliography{aaai2027}

\end{document}